\documentclass[a4paper,fleqn]{cas-dc}

\usepackage[authoryear]{natbib}
\usepackage{caption}
\usepackage{romannum}

\def\tsc#1{\csdef{#1}{\textsc{\lowercase{#1}}\xspace}}
\tsc{WGM}
\tsc{QE}
\begin{document}
\let\WriteBookmarks\relax
\def\floatpagepagefraction{1}
\def\textpagefraction{.001}

\shorttitle{}    

\shortauthors{}  

\title [mode = title]{Accurate Diagnosis of Respiratory Viruses Using an Explainable Machine Learning with Mid-Infrared Biomolecular Fingerprinting of Nasopharyngeal Secretions}  




%\cormark[1]


%\ead{}




\author[1]{Wenwen Zhang}
\credit{Writing-review \& editing, Writing-Original draft, Conceptualization, Visualization, Formal analysis, Methodology, Investigation, Data curation, Validation}

\author[2]{Zhouzhuo Tang}
\credit{Investigation, Resources, Data curation}

\author[3]{Yingmei Feng}
\credit{Investigation, Resources, Writing-review \& editing, Validation}
\author[2]{Xia Yu} 
\credit{Writing-review \& editing, Investigation, Resources, Funding acquisition, Project adminstration}
\author[1]{Qi Jie Wang}
%\cormark[2]
%\ead{qjwang@ntu.edu.sg}
\credit{Conceptualization, Funding acquisition, Investigation, Resources, Supervision, Project adminstration}
\author[1]{Zhiping Lin}
%\cormark[1]
%\cortext[cor1]{Corresponding author.}
%\ead{EZPLin@ntu.edu.sg}
%\cortext[cor2]{Corresponding author.}

\credit{Conceptualization, Writing-review \& editing, Supervision, Investigation, Resources, Project adminstration}
% Footnote of the second author

% Email id of the second author
%\ead{}




\affiliation[1]{organization={School of Electrical and Electronic Engineering, Nanyang Technological University},
	city={Singapore},
	%          citysep={}, % Uncomment if no comma needed between city and postcode
	postcode={639798}, 
	country={Singapore}}
\affiliation[2]{organization={School of Instrumentation and Optoelectronic Engineering, Beihang University},
            city={Beijing},
%          citysep={}, % Uncomment if no comma needed between city and postcode
            postcode={100191}, 
            country={China}}
        
\affiliation[3]{organization={ Department of Science and Development, Beijing Youan Hospital, Capital Medical University},
	city={Beijing},
	postcode={10069}, 
	country={China}}

%\cortext[1]{Corresponding author}



\begin{abstract}
Accurate identification of respiratory viruses (RVs) is critical for outbreak control and public health. This study presents a diagnostic system that combines Attenuated Total Reflectance Fourier Transform Infrared Spectroscopy (ATR-FTIR) from nasopharyngeal secretions with an explainable Rotary Position Embedding-Sparse Attention Transformer (RoPE-SAT) model to accurately identify multiple RVs within 10 minutes. Spectral data (4000–400 cm\textsuperscript{-1}) were collected, and the bio-fingerprint region (1800–900 cm\textsuperscript{-1}) was employed for analysis. Standard normal variate (SNV) normalization and second-order derivation were applied to reduce scattering and baseline drift. Gradient-weighted class activation mapping (Grad-CAM) was employed to generate saliency maps, highlighting spectral regions most relevant to classification and enhancing the interpretability of model outputs. Two independent cohorts from Beijing Youan Hospital, processed with different viral transport media (VTMs) and drying methods, were evaluated, with one including influenza B, SARS-CoV-2, and healthy controls, and the other including mycoplasma, SARS-CoV-2, and healthy controls. The model achieved sensitivity and specificity above 94.40\% across both cohorts. By correlating model-selected infrared regions with known biomolecular signatures, we verified that the system effectively recognizes virus-specific spectral fingerprints, including lipids, Amide \Romannum{1}, Amide \Romannum{2}, Amide \Romannum{3}, nucleic acids, and carbohydrates, and leverages their weighted contributions for accurate classification.

\end{abstract}


%\item 
%\item 
%\item 

%\nocite{*}

% Keywords

\maketitle

% Main text
\section{Introduction}\label{}
Respiratory viruses (RVs) are highly contagious pathogens that primarily affect the respiratory tract \citep{b1, b2, s1}. In particular, SARS-CoV-2 exhibits rapid human-to-human transmission, which led to the COVID-19 pandemic, characterized by high mortality among the elderly and individuals with pre-existing health conditions \citep{b3, b4}. Due to overlapping clinical symptoms among RVs, accurate differentiation is critical for appropriate treatment and transmission control. This creates a need for a rapid, reusable, and environmentally friendly diagnostic system to facilitate large-scale screening and timely medical intervention. \par  
Currently, Reverse Transcription quantitative Polymerase Chain Reaction (RT-qPCR) is the gold standard for respiratory virus (RV) detection due to its high sensitivity and specificity \citep{b5}. However, it is time-consuming, labor-intensive, and requires complex procedures and specialized infrastructure, limiting its scalability in resource-limited settings. Antigen self-test kits can identify specific viruses \citep{b6}, but their sensitivity and specificity are lower, and no universal kit exists to detect multiple RVs simultaneously. \par 
Recent advances have explored both optical and non-optical biosensors for real-time RVs detection \citep{b7, b8, b9, b10}.  Non-optical detection approaches include various techniques such as nucleic acid amplification tests (NAATs), antigen testing, serological testing, CRISPR-based detection, nanomaterial-based sensors, electrochemical detection, and immunosensor technology \citep{b11, b12, b13, b14, b15}. For instance, Dong et al. developed a triple-mode homogeneous biosensor \citep{b16} that combines electrochemical, fluorescent, and colorimetric methods for the simultaneous detection of SARS-CoV-2, influenza A, and influenza B.  \par 
Optical biosensors detect RVs through signals such as light absorption, emission, and scattering, using techniques like fluorescence, surface plasmon resonance, and Raman spectroscopy \citep{b17, b15, b19, b20}. For example, Goswami et al. employed spatial light interference microscopy (SLIM) combined with deep learning for SARS-CoV-2 detection \citep{b21}, but the approach relied on synthetic data and lacked clinical validation. However, the approach relied on synthetic data and lacked clinical validation. Yang et al. integrated surface-enhanced Raman spectroscopy (SERS) with machine learning to detect 13 RV species \citep{b22}, achieving high accuracy using nanorod array substrates. However, the method depended on spiked samples, required carefully controlled preparation, and involved complex fabrication processes, limiting its scalability and clinical applicability. \par
Infrared spectroscopy is a non-invasive technique that identifies molecular vibrations through characteristic absorption bands of functional groups. Within the 1800–900 cm\textsuperscript{-1} range, biomolecules such as proteins, lipids, carbohydrates, and nucleic acids exhibit distinct patterns, forming unique molecular fingerprints \citep{b23, b24}. Integrating infrared spectra with deep learning enables accurate, automated biological analysis. For instance, Zhang et al. \citep{b25} combined ATR-FTIR of serum samples with Partial Least Squares Discriminant Analysis (PLS-DA) and Convolutional Neural Network (CNN) for SARS-CoV-2 detection. While the study demonstrates a promising ultrafast and reagent-free diagnostic strategy using ATR-FTIR spectroscopy. However, the approach is limited by invasive sampling and the lack of model interpretability.   \par
To address the aforementioned practical challenges, we developed a diagnostic system for RV identification that integrates ATR-FTIR spectra from nasopharyngeal secretions with an explainable RoPE-SAT model. The model employs gradient-weighted class activation mapping (Grad-CAM) \citep{b26} to highlight infrared spectral bands most relevant to classification. These salient regions align with characteristic biomolecular signatures such as lipids, Amide \Romannum{1}-\Romannum{3}, nucleic acids, and carbohydrates, enabling interpretable and accurate virus discrimination. Due to the continuous and peak-oriented characteristics of infrared spectra, RoPE-based relative positional encoding is employed to capture both local patterns and global trends, thereby enhancing the model’s ability to identify virus-specific absorption peaks and underlying spectral correlations. Furthermore, sparse attention reduces computational complexity by 80\% while preserving classification performance, as detailed in Supplementary Note S3. The system achieved over 94.40\% sensitivity and specificity on two independent cohorts preserved in different VTMs and subjected to distinct drying protocols, demonstrating strong generalizability across varied sample preparation conditions.
\section{Experiment and method}
The workflow of the proposed diagnostic system is shown in Scheme \ref{framework}, comprising three main stages: 1) sample preparation, 2) infrared spectral acquisition, and 3) RV identification and interpretation via the RoPE-SAT model. To investigate the impact of potential spectral shifts introduced by different VTMs and drying protocols on the diagnostic system, two independent cohorts were established to evaluate its generalizability.
\subsection{Materials}
Ultrafiltration tubes (UFC5050) with a nominal molecular weight cutoff of 50 kDa were purchased from Merck Millipore. Viral transport media VTM1 (MT0301-1) and VTM2 (11783) were obtained from Youkang Biotechnology (Beijing) Co., Ltd. and Jiangsu CoWin Biotech Co., Ltd., respectively.
\subsection{Preparation of nasopharyngeal secretion samples}
\begin{figure*}[htbp]\centering
	\captionsetup{name=Scheme}
	\includegraphics[width=17.6cm]{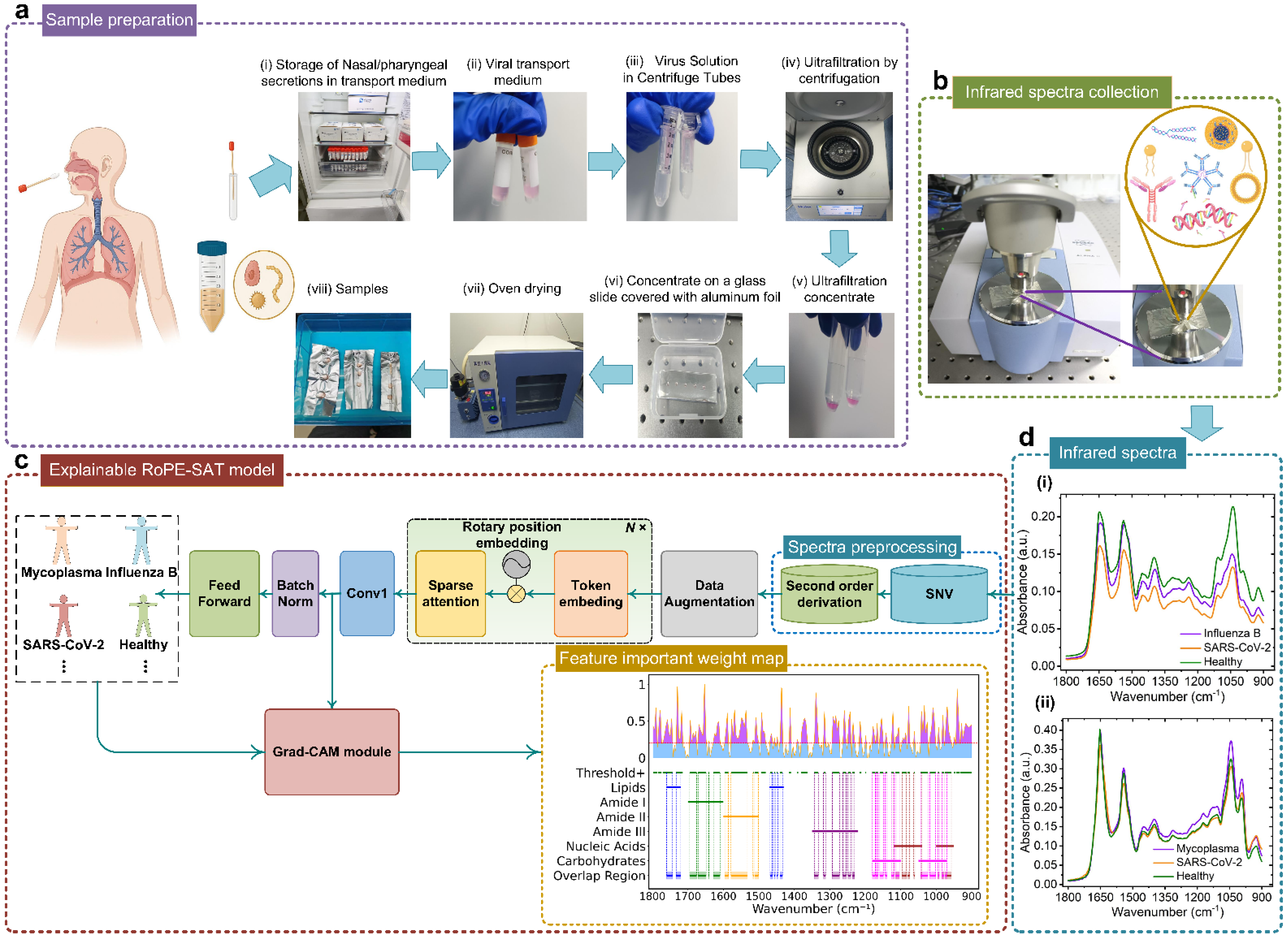}
	\caption{RVs identification using explainable RoPE-SAT model with infrared biomolecular fingerprinting of nasalpharyngeal secretions. a. Sample preparation protocol. b. Infrared spectral collection of nasalpharyngeal secretions.  c.The proposed explainable RoPE-SAT model for RV identification, with interpretability demonstrated by the overlap between model-selected infrared regions and known biomolecular absorption bands. d. Infrared spectral signals collected from two groups of nasopharyngeal swab samples preserved in different VTM solutions. (i) Influenza B, SARS-CoV-2, and healthy controls. (ii) Mycoplasma, SARS-CoV-2, and healthy controls.}
	\label{framework}
\end{figure*}

Scheme \ref{framework}a illustrates the processing of nasopharyngeal secretion samples. Swabs were collected at Beijing Youan Hospital (Table S1 and Fig. S2). Cohort 1 included patients with influenza B or SARS-CoV-2 infections and healthy controls; Cohort 2 included mycoplasma or SARS-CoV-2 cases and corresponding controls. Sample labels (Table S1 and Fig. S2) were assigned based on RT-qPCR confirmation.
Specimens were stored at –80 $^{\circ}\mathrm{C}$ in sterile tubes containing different types of VTMs, as shown in Scheme \ref{framework}a(i)-(ii). Before ATR-FTIR analysis, samples were thawed and briefly vortexed, a step that is not required for on-site screening. For both Cohorts, 400 $\mu$L of each sample (Scheme~\ref{framework}a(iii)) was concentrated and purified by ultrafiltration using a 50 kDa cutoff membrane at 14000$\times$g for 5 minutes at 4~$^{\circ}\mathrm{C}$, effectively enriching biomolecules while removing low-molecular-weight components (Scheme~\ref{framework}a(iv)). A 30 $\mu$L concentrate (Scheme~\ref{framework}a(v)) was pipetted onto aluminum foil-covered slides (Scheme~\ref{framework}a(vi)).
Cohort 1 samples were oven-dried for 2 minutes (Scheme~\ref{framework}a(vii)), while Cohort 2 samples were air-dried at room temperature for approximately 8 minutes. The final processed secretion sample is shown in Scheme~\ref{framework}a(viii).
   
\subsection{Infrared spectra collection}
As shown in Scheme~\ref{framework}b, spectral data were acquired using a Bruker Alpha \Romannum{2} FTIR spectrometer with a platinum ATR module. Measurements were performed at 1 cm\textsuperscript{-1} resolution across the 4000–400 cm\textsuperscript{-1} range (Fig. S1), with 64 scans collected for both background and sample, completing each acquisition in approximately 1 minute. The dataset comprised 423 nasopharyngeal secretion samples: 264 stored in VTM1 (96 influenza B, 72 SARS-CoV-2, 95 healthy controls) and 159 in VTM2 (46 mycoplasma, 53 SARS-CoV-2, 59 healthy controls). Each sample was measured in triplicate, and abnormal spectra were excluded, yielding 1269 spectra with 874 wavenumber points each. The biomolecular fingerprint region (1800-900 cm\textsuperscript{-1}), as illustrated in Scheme~\ref{framework}d(i–ii), was extracted for classification by the explainable RoPE-SAT model.
\begin{figure*}[htbp]\centering
	\includegraphics[width=17.6cm]{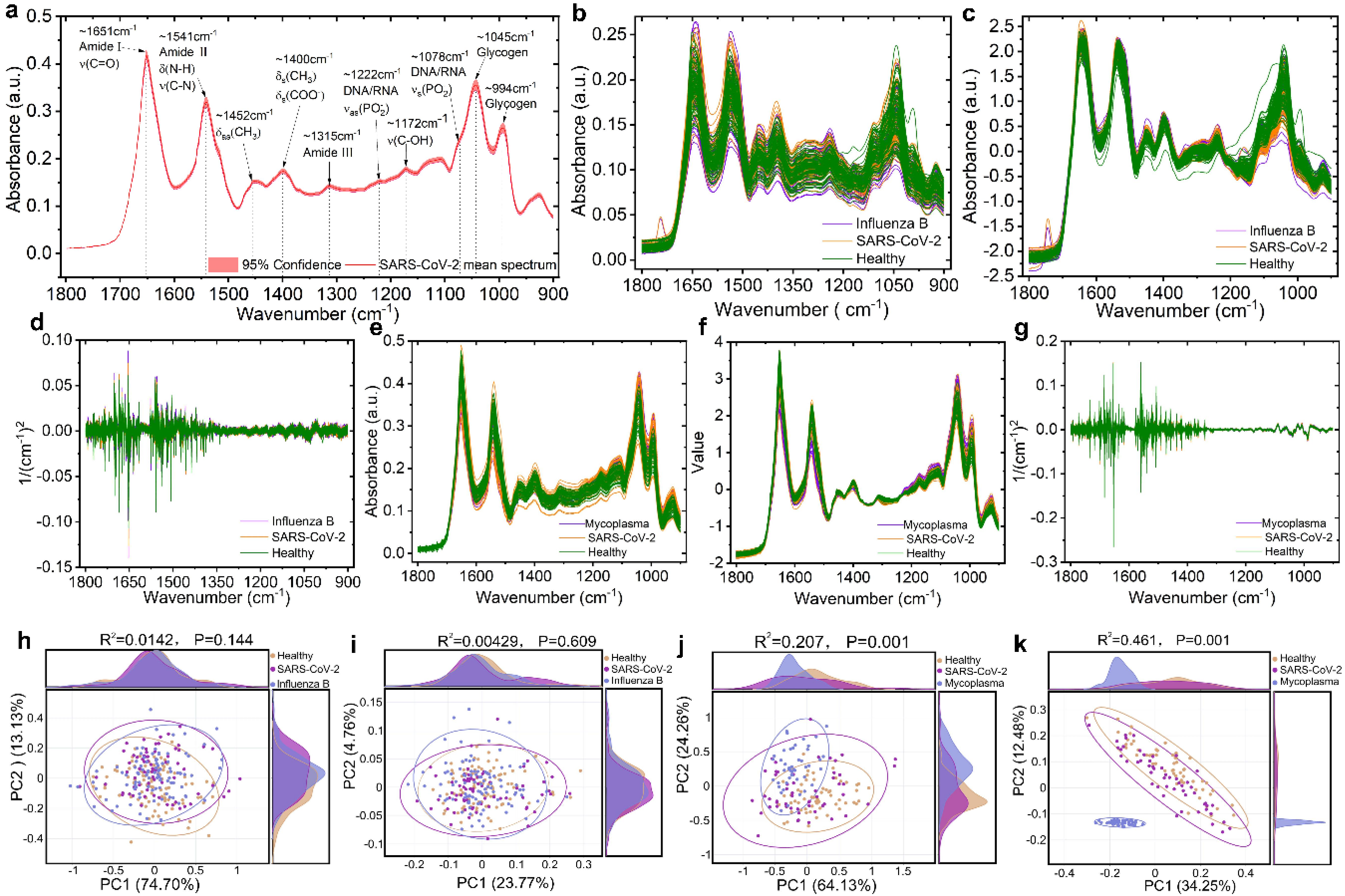}
	\caption{Characteristic absorption peaks of major biomolecules associated with viral infection, original and preprocessed spectral signals for samples from Cohort 1 and Cohort 2, and marginal PCA-based dimensionality reduction analysis before and after signal preprocessing. (a)Characteristic absorption peaks of major biomolecules associated with viral infection.  (b) Original spectral signals for cohort1. (c)Spectral signals for cohort1 after SNV preprocessing. (d)Spectral signals for cohort1 after SNV and second-order derivative preprocessing. (e)Original spectral signals for cohort2. (f)Spectral signals for cohort2 after SNV preprocessing. (g)Spectral signals for cohort2 after SNV and second-order derivative preprocessing. (h)Marginal PCA of the original spectral signals from Cohort 1. (i)Marginal PCA of SNV- and second-derivative-processed spectra from Cohort 1. (j)Marginal PCA of the original spectral signals from Cohort 2. (k)Marginal PCA of SNV- and second-derivative-processed spectra from Cohort 2.}
	\label{preprocessing}
\end{figure*}

\subsection{Spectra preprocessing}    
The raw spectral signals from influenza B, SARS-CoV-2, and healthy controls in Cohort 1, and from mycoplasma, SARS-CoV-2, and healthy controls in Cohort 2, are shown in Figs. 2b and 2e, respectively. As illustrated in Scheme~\ref{framework}c and \ref{framework}d, all spectra were first preprocessed using standard normal variate (SNV) normalization to eliminate baseline offset and scattering effects before being fed into the RoPE-SAT model.

Let the spectrum of the $i$-th sample be denoted as $S_i = {s_{i1}, s_{i2}, ..., s_{iN}}$, where $N$ is the number of wavenumber points. The mean and standard deviation of $S_i$ are calculated as: 
\begin{equation} 
\label{f1} \bar{s}_i = \frac{1}{N} \sum_{j=1}^{N}s_{ij}, \quad v_i = \sqrt{\frac{1}{N} \sum_{j=1}^{N}(s_{ij} - \bar{s}_i)^2} 
\end{equation} 
The SNV-normalized spectral value is then given by: 
\begin{equation} 
\label{f2} \hat{s}_{ij} = \frac{s_{ij} - \bar{s}_i}{v_i} 
\end{equation}

The SNV-corrected spectra of Cohort1 and Cohort2 are presented in Figs. 2c and 2f, respectively. Compared to the raw spectra in Figs. 2b and 2e, the corrected signals show improved baseline stability and enhanced fine features. Subsequently, second-order derivation was applied to further reduce noise and low-frequency interference, and to enhance spectral detail and peak contrast. The resulting signals are shown in Figs. 2d and 2g. Compared to the SNV-corrected spectra, the derivative-transformed signals exhibit clearer peak structures and improved resolution, which benefit downstream feature extraction. The second-order derivative, denoted as $\Delta^2 \hat{s}_{ij}$, is computed as:
\begin{equation} 
\label{f3} \Delta^2 \hat{s}_{ij} = \hat{s}_{i(j+1)} - 2\hat{s}_{ij} + \hat{s}_{i(j-1)} 
\end{equation} 
where $i$ and $j$ denote the indices of the sample and wavenumber, respectively. Marginal PCA-based visualizations of the original and preprocessed spectral signals from Cohort 1 are shown in Fig. \ref{preprocessing}h and i, respectively, while those from Cohort 2 are presented in Fig. \ref{preprocessing}j and k, respectively. As depicted in Fig. \ref{preprocessing}i, the preprocessed spectral signals of Cohort 1 yielded an $\rm R^2$ of only 0.00429 and a relatively high $\rm P$-value of 0.609. Moreover, the marginal density plots along both PC1 and PC2 reveal substantial overlap among the three classes: influenza B, SARS-CoV-2, and healthy, indicating limited separability in the projected feature space. This presents a significant challenge for the subsequent classification task. In contrast, as shown in Fig. \ref{preprocessing}k, the mycoplasma class in Cohort 2 is clearly distinguishable from both the SARS-CoV-2 and healthy groups after preprocessing. However, the marginal density plots along PC1 and PC2 reveal substantial overlap between the SARS-CoV-2 and healthy classes, suggesting that the primary difficulty in Cohort 2 lies in distinguishing between these two categories. 
\subsection{Spectra data augmentation}  
To further improve model generalization and increase spectral data diversity, we applied a data augmentation strategy to the preprocessed spectral signals. This strategy combines linear interpolation (Mixup), random scaling, and Gaussian noise perturbation, and is applied exclusively between samples of the same class to preserve label consistency. Let $\Delta^2\hat{S}_i = \{\Delta^2\hat{s}_{i1}, \Delta^2\hat{s}_{i2}, ..., \Delta^2\hat{s}_{iN} \}$ and $\Delta^2\hat{S}_k = \{\Delta^2\hat{s}_{k1}, \Delta^2\hat{s}_{k2}, ..., \Delta^2\hat{s}_{kN} \}$ represent two SNV- and second-derivative-processed spectra from the same class. The augmented signal $\tilde{S}_i = \{\tilde{s}_{i1}, \tilde{s}_{i2}, ..., \tilde{s}_{iN} \}$ is computed as: 
\begin{equation} 
\tilde{s}_{ij} = \alpha \cdot \left( \lambda \cdot \Delta^2\hat{s}_{ij} + (1 - \lambda) \cdot \Delta^2\hat{s}_{kj} \right) + \epsilon_j 
\end{equation} where the Mixup coefficient $\lambda \sim \text{Beta}(0.4, 0.4)$, the scaling factor $\alpha \sim \mathcal{U}(0.9, 1.1)$, and the noise term $\epsilon_j \sim \mathcal{N}(0, 0.05^2)$. During training, each sample was augmented 200 times, leading to approximately a 200-fold expansion of the training set.
\subsection{RoPE-SAT Model for spectral identification}
\begin{figure*}[htbp]\centering
	\includegraphics[width=17.6cm]{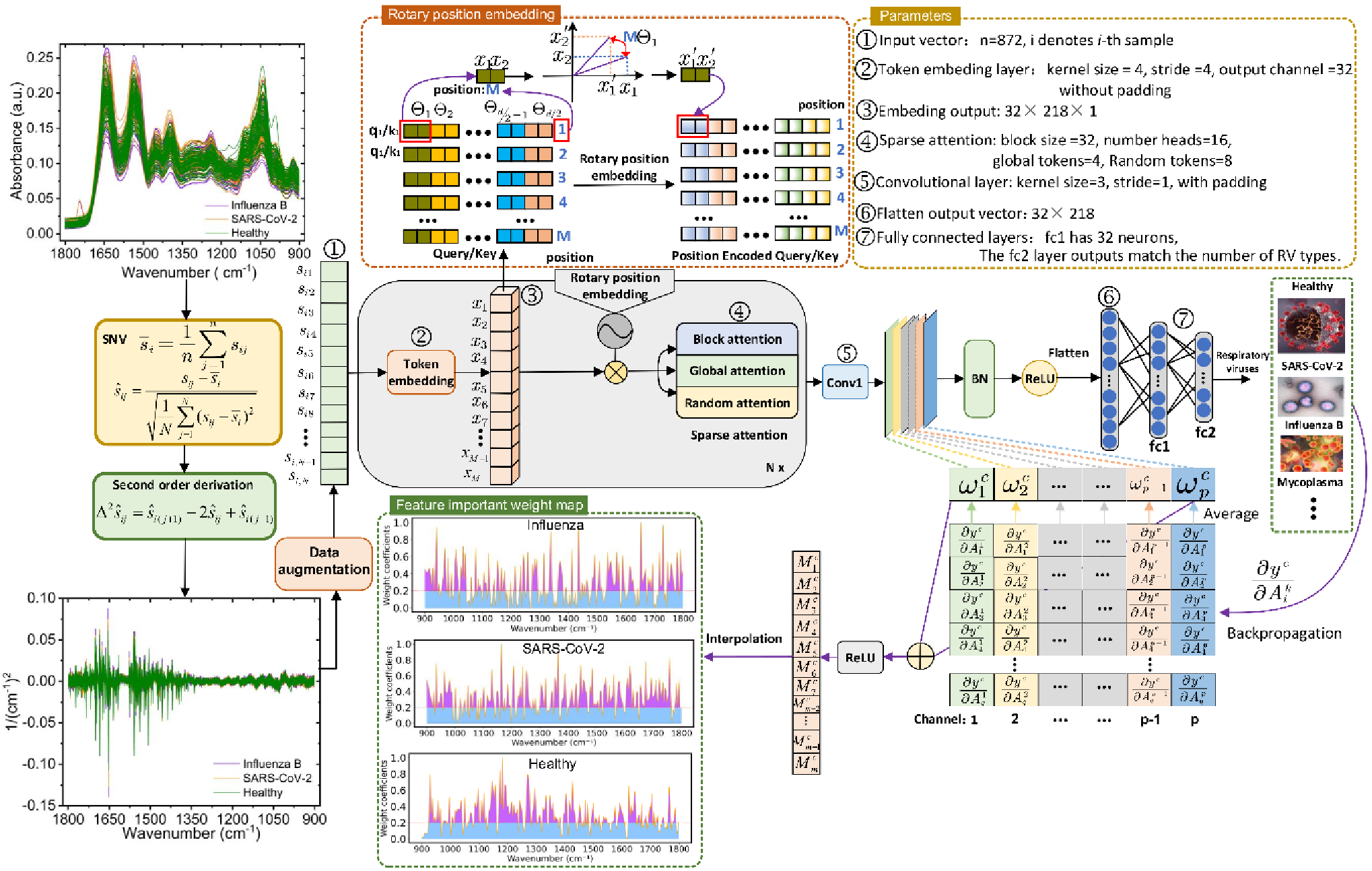}
	\caption{The detailed architecture of the proposed explainable RoPE-SAT model for spectral identification.}
	\label{model}
\end{figure*}
The detailed architecture of the proposed explainable RoPE-SAT model, designed to distinguish influenza B, SARS-CoV-2, and healthy samples from Cohort 1 (preserved in VTM1), is illustrated in Fig. 3. The acquired spectral signals are randomly shuffled and split into 80\% training and 20\% test sets. Data augmentation is applied to the training set, with the augmented spectral signals treated as ordered sequences. These sequences are first processed by a convolutional embedding layer with a kernel size of 4 and stride of 4, which downsamples the input while preserving key spectral features. Serving as the token embedding module, this layer projects the input into a latent space, capturing local spectral dependencies. To encode positional information efficiently, the model integrates RoPE into the self-attention mechanism. Unlike traditional absolute position encoding, RoPE preserves relative positional relationships, enhancing pattern recognition and translation invariance (Supplementary Note S2). The model further incorporates a sparse attention mechanism comprising block, global, and random attention modules. This structure selectively focuses on informative spectral regions, improving efficiency while retaining key discriminative features (Supplementary Note S3). The refined attention features are passed through a convolutional layer, followed by batch normalization and ReLU activation, to enhance representation. The resulting high-dimensional feature vector is flattened and passed through fully connected layers, culminating in a classification layer that predicts the respiratory virus type: SARS-CoV-2, influenza B or healthy. The architectural details of the RoPE-SAT model are presented in Supplementary Note S1, while a comprehensive description of the Grad-CAM methodology is provided in Supplementary Note S4. Detailed layer-wise configuration of the RoPE-SAT model is provided in Table S9.

The same experimental procedure is applied to distinguish mycoplasma, SARS-CoV-2, and healthy samples in Cohort 2 (preserved in VTM2). All experiments employ five-fold cross-validation. The accuracy and loss curves across training epochs for Cohort 1 and Cohort 2 are shown in Fig. \ref{Influ}e–f and Fig. \ref{mycop}e–f, respectively. After cross-validation, the model achieved average classification accuracies of 95.82\% for Cohort 1 and 96.86\% for Cohort 2.
\section{Results and discussion}
\subsection{Model Explanation for Classifying Cohort 1 and discussion}
\begin{figure*}[htbp]\centering
	\includegraphics[width=17.6cm]{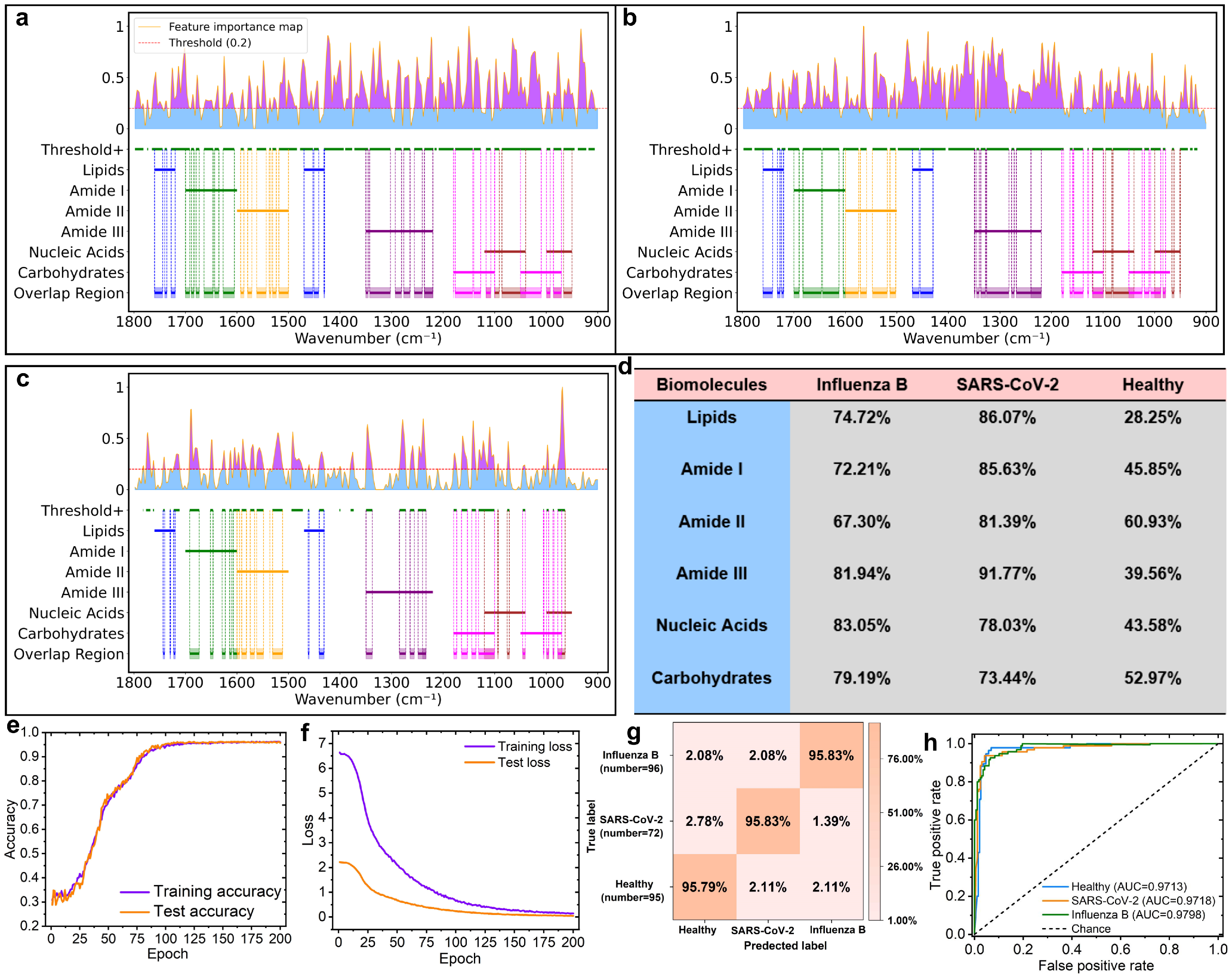}
	\caption{Determination of the overlap ratio between model-selected salient infrared regions and known biomolecular absorption bands using feature importance maps generated by the RoPE-SAT model for classifying influenza B, SARS-CoV-2, and healthy controls. 
		(a)–(c)Overlapping regions between salient infrared ranges (feature importance weight > 0.2) selected by the model and known biomolecular absorption bands, as visualized through feature importance weight maps for (a)influenza B, (b)SARS-CoV-2, and (c)healthy controls. (d)Overlap ratios for each biomolecule. (e)Accuracy curves. (f)Loss curves. (g)Confusion matrix of the classification results. (h)ROC curves and corresponding AUC values.   }
	\label{Influ}
\end{figure*}

Conventional deep learning models are often treated as black boxes, evaluated primarily based on classification accuracy, with limited interpretability \citep{b27, b28, b29}. To address this, we draw inspiration from Grad-CAM \citep{b30}, which highlights critical input regions that a CNN focuses on during classification. We propose an approach to interpret the decision-making process of the RoPE-SAT model by calculating the overlap ratio between the salient infrared regions identified by Grad-CAM and the known absorption bands of biomolecules.

During training, Grad-CAM extracts gradient information from the final convolutional layer to indicate each feature's contribution to virus identification. A feature importance weight map is then constructed by applying gradient-based weights to the layer’s activations, followed by ReLU activation. Fig. \ref{model} illustrates this process. To determine the model’s salient infrared regions, we define a weight threshold $\beta$, with features exceeding this threshold considered important. %These regions are then compared with the characteristic absorption bands of key biomolecules and chemical functional groups, such as lipids, Amide \Romannum{1} through \Romannum{3}, nucleic acids, and carbohydrates, which constitute viral structures. The RoPE SAT model leverages these spectral distinctions to classify respiratory viruses.%

To quantify the correlation between model-identified salient regions and known biomolecular bands, we introduce a metric, $\gamma$, termed the overlap ratio: \begin{equation} \gamma = \frac{\sum_{\hat{w} \in \text{overlap region}} \hat{w}}{\sum_{w \in \text{biomolecule ranges}} w} \end{equation} Here, the numerator sums the weights of overlapping spectral bands with importance values exceeding 0.2 (as shown in Fig. 4a–c), and the denominator represents the total absorption range of relevant biomolecules. A higher $\gamma$ indicates a greater contribution of biomolecular signatures to model predictions.

Fig. 4a–c illustrate the overlap between model-selected salient infrared ranges (feature importance > 0.2) and known biomolecular absorption bands, derived from the feature importance map for classifying influenza B, SARS-CoV-2, and healthy controls. Fig. 4d presents the overlap ratio for each biomolecule across these classifications. A key observation is the significantly higher lipid overlap in SARS-CoV-2 (86.07\%) and influenza B (74.72\%) compared to healthy individuals (28.25\%), indicating virus-induced lipid metabolism remodeling. Lipid membranes are essential for viral entry, replication, and immune evasion. As an enveloped RNA virus, SARS-CoV-2 extensively remodels host lipid rafts and phospholipids to facilitate endocytosis and intracellular trafficking \citep{b31}. Similarly, influenza B exploits host lipid droplets for genome replication and virion assembly, albeit to a lesser extent than SARS-CoV-2 \citep{b32}. In contrast, the lower lipid overlap observed in healthy individuals reflects a stable lipid profile in the absence of virus-induced disruptions.
Nucleic acids also play a key role, with spectral overlap rates of 83.05\% for influenza B and 78.03\% for SARS-CoV-2, both notably higher than the 43.58\% observed in healthy individuals. This indicates that viral infections substantially alter host nucleic acid metabolism, consistent with the fact that RNA viruses rely on extensive replication within host cells \citep{b33}. As RNA viruses, influenza B and SARS-CoV-2 depend on the host's cellular machinery for genome replication, resulting in increased viral RNA accumulation \citep{b34}. In contrast, the lower spectral overlap observed in healthy controls reflects baseline nucleic acid levels in the absence of viral interference.  The overlap rates for Amide I, II, and III bands show a clear trend of increased spectral similarity in infected cases, consistent with virus-induced alterations in host protein structures. Amide I ($\sim$1650 cm\textsuperscript{-1}), associated with C=O stretching in $\alpha$-helices and $\beta$-sheets, shows overlap rates of 85.63\% for SARS-CoV-2, 72.21\% for influenza B, and only 45.83\% in healthy individuals, indicating significant changes in protein secondary structure due to viral infection \citep{b35}. Amide II ($\sim$1540 cm\textsuperscript{-1}), arising from N–H bending vibrations and related to protein backbone conformation, shows similar trends, suggesting disruptions in translation and immune-related synthesis \citep{b36}. Amide III ($\sim$1235 cm\textsuperscript{-1}), associated with C–N stretching and N–H bending vibrations linked to protein folding and tertiary structure, exhibits the highest overlap in SARS-CoV-2 cases at 91.77\%, followed by influenza B at 81.94\% and healthy individuals at 39.56\%, which is closely related to significant conformational protein rearrangements induced by viral infection \citep {b37}. These findings collectively indicate that our model effectively identifies the substantial impact of viral infections, especially SARS-CoV-2, on protein secondary and tertiary structures, attributed to enhanced viral protein synthesis, host translation disruption, and stress-induced misfolding. Moreover, Carbohydrate overlap rates of 77.50\% for influenza B, 71.88\% for SARS-CoV-2, and 51.88\% for healthy controls support the concept that viral infections perturb glycosylation pathways and carbohydrate metabolism \citep{b38}. In contrast, the lower overlap in healthy individuals reflects the absence of viral glycoprotein production.\par    
\begin{figure*}[htbp]\centering
	\includegraphics[width=17.6cm]{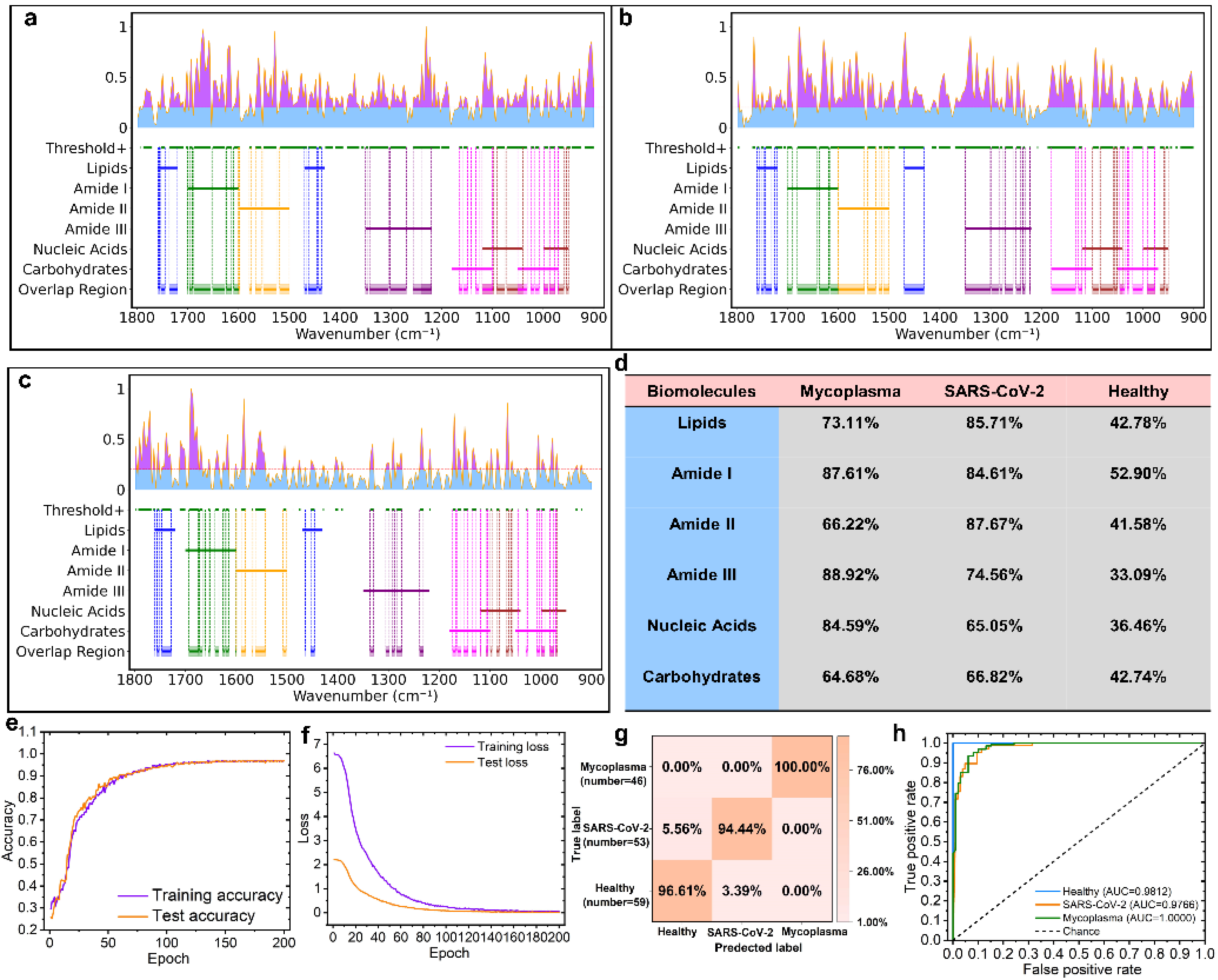}
	\caption{Determination of the overlap ratio between model-selected salient infrared regions and known biomolecular absorption bands using feature importance maps generated by the RoPE-SAT model for classifying mycoplasma, SARS-CoV-2, and healthy controls. 
		(a)–(c)Overlapping regions between salient infrared ranges (feature importance weight > 0.2) selected by the model and known biomolecular absorption bands for (a)mycoplasma, (b)SARS-CoV-2, and (c)healthy controls. (d)Overlap ratios for each biomolecule. (e)Accuracy cureves. (f)Loss curves. (g)Confusion matrix of the classification results. (h)ROC curves and corresponding AUC values.   }
	\label{mycop}
\end{figure*}

\subsection{Model Explanation for Classifying Cohort 2 and discussion}
Fig. \ref{mycop}a to \ref{mycop}c illustrate the overlap between model-identified salient infrared regions and known biomolecular absorption bands, derived from the feature importance weight map for classifying mycoplasma, SARS CoV 2, and healthy controls. Fig. \ref{mycop}d presents the overlap ratios for each biomolecule across these categories. The higher lipid overlap observed in SARS-CoV-2 (85.71\%) and mycoplasma (71.25\%), compared to healthy controls (42.78\%), reflects pathogen induced remodeling of host lipid metabolism. This is consistent with the fact that SARS CoV 2 actively rewires lipid pathways by reorganizing lipid rafts and upregulating phospholipid biosynthesis to facilitate viral entry, replication organelle formation, and envelope acquisition \citep{b39}. In contrast, mycoplasma, lacking a cell wall and possessing limited biosynthetic capacity, depends on scavenging host derived lipids such as cholesterol and phosphatidylcholine to maintain membrane integrity \citep{b40, b41}. These fundamental differences in lipid utilization are reflected in the model's distinct spectral response profiles.

Similarly, the greater nucleic acid overlap observed in mycoplasma (84.59\%), compared to SARS-CoV-2 (65.05\%) and healthy controls (36.46\%), can be attributed to mycoplasma's reduced genome and its inability to synthesize nucleotides, which forces it to rely on host derived nucleic acids \citep{b42}. This results in stronger absorption signals in nucleic acid associated spectral regions, particularly around 1080 to 1120 wavenumbers, as shown in Fig. \ref{preprocessing}f. In contrast, SARS-CoV-2 primarily targets host messenger RNA for degradation while sparing nuclear DNA, and healthy controls display a stable nucleic acid profile without pathogenic disruption. These differences suggest that increasing the model’s sensitivity to the nucleic acid region can enhance its ability to discriminate mycoplasma infections from influenza B.

The increased overlap ratios observed in mycoplasma and SARS-CoV-2 infections across the Amide \Romannum{1}-\Romannum{3} regions suggest pronounced protein structural alterations associated with pathogenic processes. Notably, the significantly elevated overlap in the Amide \Romannum{2} band for SARS-CoV-2 samples likely reflects enhanced viral protein synthesis and the translational hijacking of host cellular machinery during infection \citep{b43}. In contrast, mycoplasma shows minimal protein expression and reduced translation due to its streamlined genome and limited regulatory capacity \citep{b44}. Consistent with the overlap patterns observed in Cohort 1, the model assigns higher importance to the carbohydrate spectral region when identifying both SARS-CoV-2 and Mycoplasma infections, compared to healthy individuals. \par
In summary, these findings demonstrate that the model effectively captures biologically meaningful variations in host–pathogen interactions by identifying distinct infrared spectral signatures, thereby enabling accurate differentiation among respiratory infections.
\begin{figure*}[htbp]\centering
	\includegraphics[width=17.0cm]{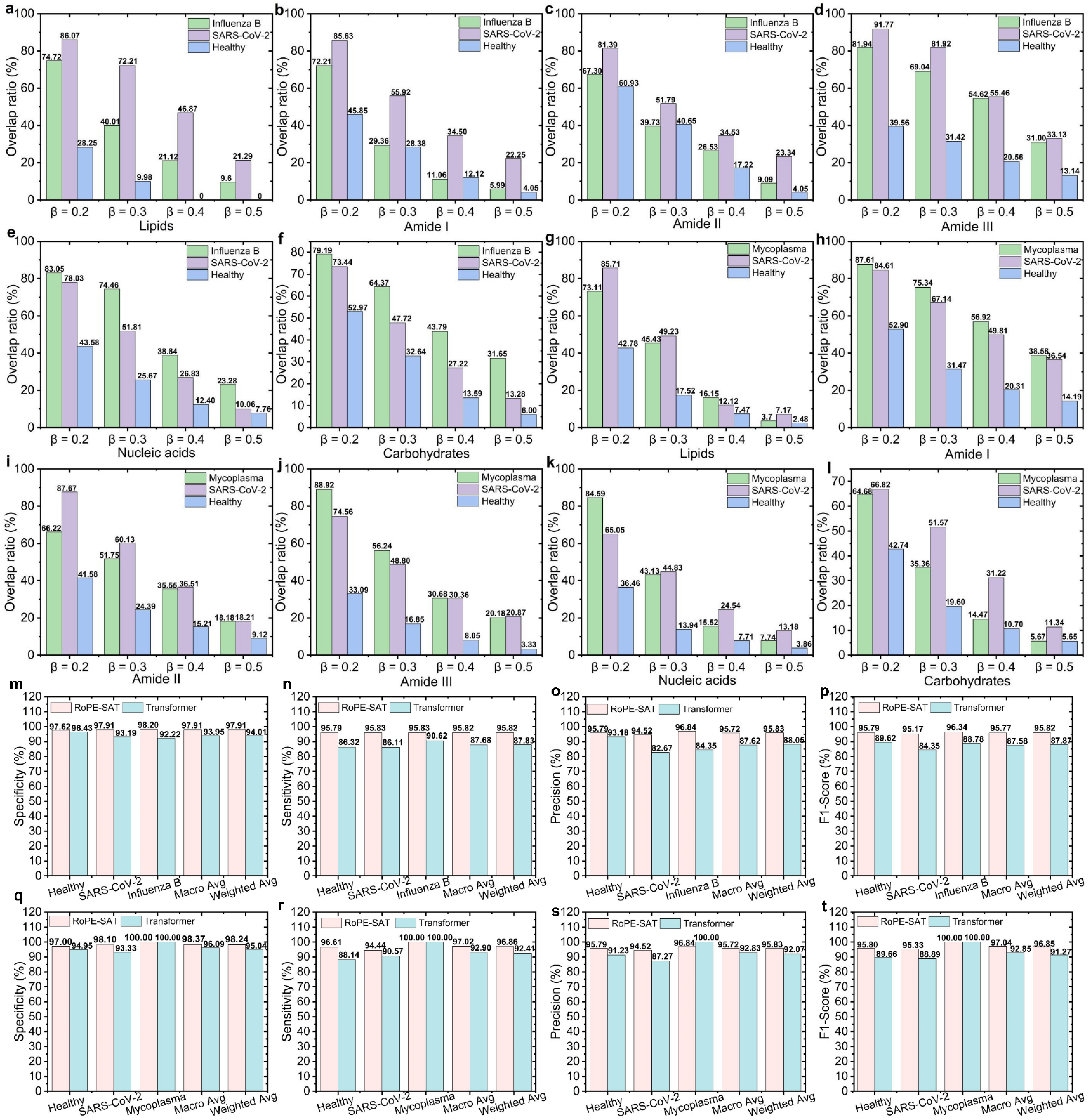}
	\caption{Overlap ratios obtained under thresholds $\beta = 0.2$, $0.3$, $0.4$, and $0.5$ and comparison of experimental results between the RoPE-SAT and the Transformer model after five-fold cross-validation. (a)Overlap ratio with lipid regions for Cohort 1. (b)Overlap ratio with Amide \Romannum{1} regions for Cohort 1. (c)Overlap ratio with Amide \Romannum{2} regions for Cohort 1. (d) Overlap ratio with Amide \Romannum{3} regions for Cohort 1. (e) Overlap ratio with nucleic acids regions for Cohort 1. (f) Overlap ratio with carbohydrates regions for Cohort 1. (g)Overlap ratio with lipid regions for Cohort 2. (h)Overlap ratio with Amide \Romannum{1} regions for Cohort 2. (i)Overlap ratio with Amide \Romannum{2} regions for Cohort 2. (j)Overlap ratio with Amide \Romannum{3} regions for Cohort 2. (k)Overlap ratio with nucleic acids regions for Cohort 2. (l)Overlap ratio with carbohydrates regions for Cohort 2. (m)Specificity of Cohort 1. (n)Sensitivity of Cohort 1. (o)Precision of Cohort 1. (p)F1-Score of Cohort 1. (q)Specificity of Cohort 2. (r)Sensitivity of Cohort 2. (s)Precision of Cohort 2. (t)F1-Score of Cohort 2.}
	\label{comparison}
\end{figure*}
\subsection{Experimental results}
To evaluate the impact of different thresholds on model interpretability, we analyzed the overlap between salient infrared spectral regions and known biomolecular absorption bands across a range of thresholds ($\beta = 0.2, 0.3, 0.4, 0.5$). These overlap regions are visualized in Fig. S3–S8, with corresponding overlap ratios summarized in Tables S3–S8. As shown in Fig. \ref{comparison}a–i, the overlap ratios for individual biomolecules exhibit consistent trends across thresholds, indicating that varying $\beta$ has minimal influence on the model's interpretation of biomolecular importance. Detailed discussions are provided in Supplementary Note S5. \par
We employed evaluation metrics including specificity, sensitivity, precision, and F1-score, and additionally utilized a Transformer model for comparison. The results are presented in Fig.~\ref{comparison}m–t.  Despite limited class separability in Cohort 1 as shown in Fig. \ref{preprocessing}i, the model achieved sensitivity of 95.79\%, 95.83\%, and 95.83\% and specificity of 97.62\%, 97.91\%, and 98.20\% for healthy, SARS-CoV-2, and influenza B, respectively. As shown in Fig. \ref{preprocessing}k for Cohort 2, the mycoplasma class forms a compact and well-separated cluster with a clearly defined confidence ellipse, and both the RoPE-SAT and Transformer models accurately identify it. In contrast, the confidence ellipses of the SARS-CoV-2 and healthy groups largely overlap, indicating limited separability. Nevertheless, the model achieves sensitivities of 96.61\% and 94.41\%, and specificities of 97.00\% and 98.10\% for healthy and SARS-CoV-2, respectively. As shown in Fig. S9, based on five-fold cross-validation, the proposed RoPE-SAT model achieved classification accuracies of 95.82\% and 96.86\% on Cohort 1 and Cohort 2, respectively, outperforming the Transformer model, which achieved 87.83\% and 92.41\%. The confusion matrices and ROC curves for Cohort 1 and Cohort 2 are presented in Fig. \ref{Influ}g–h and Fig. \ref{mycop}g–h, respectively. The model exhibits a tendency to misclassify SARS-CoV-2 samples as healthy in both Cohort 1 and Cohort 2. Compared to the Transformer model results shown in Fig. S10a and S10b, the proposed model achieves consistently higher AUC scores across both Cohort 1 and Cohort 2.    

\section{Conclusion and future work}
In this study, we developed a diagnostic system for respiratory viruses that integrates ATR-FTIR spectral signals from nasopharyngeal secretions with the explainable RoPE-SAT deep learning model, enabling highly accurate identification within 10 minutes. By calculating the overlap between the major infrared absorption regions of key viral molecular components and the salient spectral bands selected by the model, we provide interpretability for the model’s outcomes. High and consistent performance was achieved in both cohorts using different viral transport media and drying protocols, with specificity exceeding 97.62\% and sensitivity reaching 95.79\% in Cohort 1, and specificity over 97.00\% and sensitivity above 94.44\% in Cohort 2, demonstrating the effectiveness and robustness of the proposed diagnostic system. The proposed RoPE-SAT model achieves approximately an 80\% reduction in attention computation while maintaining high sensitivity and specificity. Moreover, Marginal PCA analysis conducted on mycoplasma samples before and after preprocessing demonstrates that the proposed diagnostic system  can reliably distinguish between respiratory  bacterial pathogens and RNA viruses. the results also validate that the applied signal preprocessing techniques, including SNV normalization and second-order derivation, effectively enhance the separability of the data. The proposed diagnostic system is expected to serve as a valuable, reproducible, cost-effective tool for large-scale on-site screening of respiratory viral infections. 
In future work, we plan to collect a broader range of respiratory virus samples to further validate the specificity and sensitivity of the proposed diagnostic system, and the interpretability of the underlying model.

% Use the style of numbering in square brackets.
% If nothing is used, default style will be taken.
%\begin{enumerate}[a)]
%\item 
%\item 
%\item 
%\end{enumerate}  

%\begin{itemize}
%\item 
%\item 
%\item 
%\end{itemize}  

%\begin{description}
%\item[]
%\item[] 
%\item[] 
%\end{description}  

% Figure

% Uncomment and use as the case may be
%\begin{theorem} 
%\end{theorem}

% Uncomment and use as the case may be
%\begin{lemma} 
%\end{lemma}


\printcredits 
\section*{Declaration of competing interest}
The authors declare that they have no known competing financial interests or personal relationships that could have appeared to influence the work reported in this paper.
\section*{Data availability}
Data will be made available on request.
\section*{Acknowledgements}
This work was partially supported by the National Key R\&D Program of China (Grant No. 2022YFE0102300), the National Medical Research Council (NMRC) under the Singapore–China Joint Grant on Infectious Diseases (MOH-000927), A*STAR grants (A2090b0144, M22K2c0080, and R23I0IR041), National Research Foundation (NRF) grant (NRF-CRP29-2022-0003), and the National Natural Science Foundation of China (Grant No. 62203307).

\bibliographystyle{cas-model2-names}

\bibliography{cas-refs}



\end{document}